\pdfoutput=1

\documentclass[11pt]{article}

\usepackage{EMNLP2023}

\usepackage{times}
\usepackage{latexsym}

\usepackage[T1]{fontenc}

\usepackage[utf8]{inputenc}

\usepackage{microtype}

\usepackage{inconsolata}

\usepackage{amsmath}
\usepackage{amsfonts}

\usepackage[T1]{fontenc}    
\usepackage{url}            
\usepackage{booktabs, multirow, tabularx}       
\usepackage{amsfonts}       
\usepackage{nicefrac}       
\usepackage{microtype}      
\usepackage{graphicx}
\usepackage{natbib}
\usepackage{doi}
\usepackage{color}
\usepackage{amsmath}
\usepackage{mathtools, cuted}
\usepackage{etoolbox}
\usepackage{changepage}
\usepackage{pdflscape}
\usepackage{multicol}
\usepackage{balance}
\usepackage{wrapfig}
\usepackage{array}
\usepackage{soul}
\usepackage{bm}
\usepackage{enumitem}
\DeclarePairedDelimiter\ceil{\lceil}{\rceil}
\DeclarePairedDelimiter\floor{\lfloor}{\rfloor}

\AfterEndEnvironment{strip}{\leavevmode}

\newcommand{\nce}{\mathrm{NCE}}

\newcommand{\JBert}{\texttt{Jina\,BERT}}
\newcommand{\AdaTwo}{\texttt{text-embedding-ada-002}}
\newcommand{\JEmbeddingVOne}{\texttt{Jina\,\,Embeddings\,v1}}
\newcommand{\JEmbeddingVTwo}{\texttt{Jina\,\,Embeddings\,v2}}

\newcommand{\JinaSTwo}{\href{https://huggingface.co/jinaai/jina-embedding-s-en-v2}{\texttt{jina-small-v2}}}
\newcommand{\JinaBTwo}{\href{https://huggingface.co/jinaai/jina-embedding-b-en-v2}{\texttt{jina-base-v2}}}

\newcommand{\rom}[1]{\uppercase\expandafter{\romannumeral #1\relax}}

\title{\textsc{Jina Embeddings\,2}: $8192$-Token General-Purpose Text Embeddings for Long Documents}
\author{Michael G\"unther, Jackmin Ong, Isabelle Mohr,  Alaeddine Abdessalem, \textbf{Tanguy Abel}, \\ \textbf{Mohammad Kalim Akram}, \textbf{Susana Guzman}, \textbf{Georgios Mastrapas}, \textbf{Saba Sturua}, \\ \textbf{Bo Wang}, \textbf{Maximilian Werk}, \textbf{Nan Wang} \and \textbf{Han Xiao}
\\
	Jina AI GmbH, Ohlauer Str. 43, 10999 Berlin, Germany \\
	\texttt{\{michael.guenther, jackmin.ong, isabelle.mohr alaeddine.abdessalem,} \\
\texttt{tanguy.abel, kalim.akram, susana.guzman, georgios.mastrapas, saba.sturua,} \\
\texttt{bo.wang, maximilian.werk, nan.wang, han.xiao\}@jina.ai}}

\date{2023/10/31}

\hypersetup{
pdftitle={Jina Embeddings 2: 8192-Token General-Purpose Text Embeddings for Long Documents},
pdfauthor={Michael G\"unther, Jackmin Ong, Isabelle Mohr,  Alaeddine Abdessalem, Tanguy Abel, Mohammad Kalim Akram, Susana Guzman, Georgios Mastrapas, Saba Sturuam, Bo Wang, Maximilian Werk, Nan Wang, Han Xiao},
pdfkeywords={Embeddings, Large Language Models, LLM, Training, 8K, OpenAI, BERT, Token length},
}

\begin{document}
\maketitle
\begin{abstract}
Text embedding models have emerged as powerful tools for transforming sentences into fixed-sized feature vectors that encapsulate semantic information. While these models are essential for tasks like information retrieval, semantic clustering, and text re-ranking, most existing open-source models, especially those built on architectures like BERT, struggle to represent lengthy documents and often resort to truncation. One common approach to mitigate this challenge involves splitting documents into smaller paragraphs for embedding. However, this strategy results in a much larger set of vectors, consequently leading to increased memory consumption and computationally intensive vector searches with elevated latency.

To address these challenges, we introduce \JEmbeddingVTwo, an open-source text embedding model\footnote{Base model (0.27G): \url{https://huggingface.co/jinaai/jina-embeddings-v2-base-en}\\Small model (0.07G): \url{https://huggingface.co/jinaai/jina-embeddings-v2-small-en}\\API: \url{https://jina.ai/embeddings}} capable of accommodating up to $8192$ tokens. This model is designed to transcend the conventional $512$-token limit and adeptly process long documents. \JEmbeddingVTwo{} not only achieves state-of-the-art performance on a range of embedding-related tasks in the MTEB benchmark but also matches the performance of OpenAI's proprietary \AdaTwo~model. Additionally, our experiments indicate that an extended context can enhance performance in tasks such as NarrativeQA.

\end{abstract}

\section{Introduction}
\label{sec:introduction}

Using neural networks to encode text and images into embedding representations has become a standard practice for analyzing and processing vast amounts of unstructured data. In natural language processing, sentence embedding models~\cite{reimers2019sentence} transform the semantics of phrases, sentences, and paragraphs into points within a continuous vector space. These transformed data points can subsequently be used for a myriad of downstream applications, such as information retrieval, as well as clustering and classification tasks.

Despite the numerous applications of embedding models, a prevailing challenge faced by many models is the limitation on the maximum sequence lengths of text that can be encoded into a single embedding. To circumvent this, practitioners often segment documents into smaller chunks prior to encoding. This tactic, unfortunately, results in fragmented semantic meanings, causing the embeddings to misrepresent the entirety of paragraphs. Furthermore, this method yields a plethora of vectors, culminating in heightened memory usage, increased computational demands during vector searches, and extended latencies. The dilemma is exacerbated when embedding vectors are stored in database systems that construct memory-intensive index structures.

The root of these text length restrictions can be traced back to the BERT architecture, which underpins most of the current open-source models. The authors of~\cite{press2022alibi} demonstrated that these models struggle to accurately represent long documents. They introduced an alternative positional embedding method named ALiBi, enabling efficient training of models to encode long text sequences. Regrettably, up until this point, the approach was exclusively employed for generative language models, neglecting its potential for open-source encoder language models aimed at crafting document embeddings. This research bridges that gap by incorporating ALiBi bidirectionally into the BERT framework, rendering it apt for encoding tasks. As a result, it empowers users to utilize it for downstream operations on texts spanning up to $8192$ tokens. Moreover, we fine-tuned this enhanced BERT model, harnessing hundreds of millions of text samples to encode texts into singular embedding representations. Our model's resultant embeddings outshine those of the \JEmbeddingVOne{} model suite~\cite{gunther2023jina} in the MTEB benchmark and rival the prowess of state-of-the-art models like E5~\cite{wang2022text}. We also found that large context lengths can amplify the efficacy of numerous downstream tasks tied to embeddings. Given that the majority of available embedding evaluation datasets comprise mainly brief text passages, we have curated datasets encompassing long text values to better evaluate embeddings. These datasets, alongside our models, are made accessible via our Hugging Face repository\footnote{\url{https://huggingface.co/jinaai}}.

This paper is structured as follows: We begin with an overview of related work in Section~\ref{sec:related_work}. This is followed by an outline of the training paradigm in Section~\ref{sec:training-overview}, a description of the backbone model and its pre-training in Section~\ref{sec:backbone_pretraining}, and a detailed walkthrough of the fine-tuning process for embeddings generation in Section~\ref{sec:embedding-training}. We culminate with an exhaustive evaluation in Section~\ref{sec:evaluation} and conclusions in Section~\ref{sec:conclusion}.

\section{Related Work}
\label{sec:related_work}

Embedding training has undergone significant evolution, transitioning from foundational techniques such as Latent Semantic Indexing (LSA) \cite{deerwester} and Latent Dirichlet Allocation (LDA) \cite{NIPS2001_296472c9} to the sophisticated prowess of pre-trained models like Sentence-BERT \cite{reimers2019sentence}. A notable shift in recent advancements is the emphasis on unsupervised contrastive learning, as showcased by works like \cite{gao2022simcse, wang2022text}. Pioneering models like Condenser \cite{gao2021condenser} and RetroMAE \cite{xiao2022retromae} have brought forth specialized architectures and pre-training methods explicitly designed for dense encoding and retrieval.

The E5~\cite{wang2022text}, \JEmbeddingVOne~\cite{gunther2023jina}, and GTE~\cite{li2023general} collections of embedding models represent another leap forward. These models propose a holistic framework tailored for effective training across a myriad of tasks. This framework adopts a multi-stage contrastive training approach. An initial phase focuses on training using a vast collection of weak pairs sourced from public data, enhancing the model's domain generalization. Following this, a supervised fine-tuning stage employs a curated set of annotated text triples, representing diverse tasks. Together, these sequential stages yield state-of-the-art outcomes on the MTEB benchmark.

Yet, despite such advancements, a glaring limitation persists: the $512$-token constraint on input sequences, stemming from foundational models like BERT. This cap is insufficient for encoding lengthy documents, often exceeding a page. ALiBi~\cite{press2022alibi} emerges as a promising solution, presenting a technique that sidesteps conventional positional embeddings and facilitates training on sequences exceeding $2048$ tokens. Notably, its typical application is centered around generative models, which inherently adopt a unidirectional bias, rendering it less suitable for embedding tasks.

Effective evaluation remains paramount for embedding models, ensuring they meet the diverse demands of real-world applications. The BEIR benchmark \cite{thakur2021beir} stands out, offering evaluations across a set of retrieval tasks and settings. Similarly, the MTEB benchmark \cite{muennighoff2023mteb} highlights the extensive applicability of text embeddings, spanning a variety of tasks and languages. However, a notable gap in both benchmarks is their limited focus on encoding long documents --- a critical aspect for comprehensive embedding evaluation.







\section{Training Paradigm Overview}
\label{sec:training-overview}

The training paradigm for \JEmbeddingVTwo~is divided into three stages:

\begin{enumerate}[label=\Roman*]
\item \textbf{Pre-training a Modified BERT:} For the backbone language model, we propose a modified BERT model capable of encoding documents with up to $8192$ tokens. This model is trained from scratch on a full-text corpus using a masked language modeling objective.

\item \textbf{Fine-tuning with Text Pairs:} To encode a text passage into a single vector representation, the model is fine-tuned on text pairs.\label{stage:two}

\item \textbf{Fine-tuning with Hard Negatives:} The model is further fine-tuned using text pairs complemented with hard negatives. This stage is crucial for enabling the model to better distinguish between relevant passages and related, but irrelevant text passages.\label{stage:three}
\end{enumerate}

While both stages~\ref{stage:two} and~\ref{stage:three} are geared towards training the models for embedding tasks, the latter is especially critical for improving the model's performance in retrieval and classification tasks (refer to Section~\ref{sec:evaluation_jina_embedding}).

\section{Pre-training a Modified BERT}
\label{sec:backbone_pretraining}

For the backbone language model, we introduce a novel transformer based on BERT~\cite{devlin2019bert} with several modifications to enhance its ability to encode extended text sequences and to generally bolster its language modeling capabilities. For the training process, we largely adopt the approach described in~\cite{liu2019roberta}, incorporating additional performance optimizations.

\subsection{Model Architecture}

\begin{table}
\begin{tabular}{ l|rrr  }
 \toprule
 Model & Layers & Hidden & Params \\
 \midrule
\JBert~Small & 4 & 512 & 33M \\
\JBert~Base & 12 & 768 & 137M \\
\JBert~Large & 24 & 1024 & 455M \\
 \bottomrule
\end{tabular}
\caption{Architecture specifications for the \JBert~models of varying sizes. The number of attention heads is selected to ensure a consistent head dimension of $64$.}
\end{table}

\paragraph{Attention with Linear Biases:}

\begin{figure*}[htb!]
\begin{minipage}[t]{0.5\textwidth}
    \centering
    \includegraphics[width=8cm]{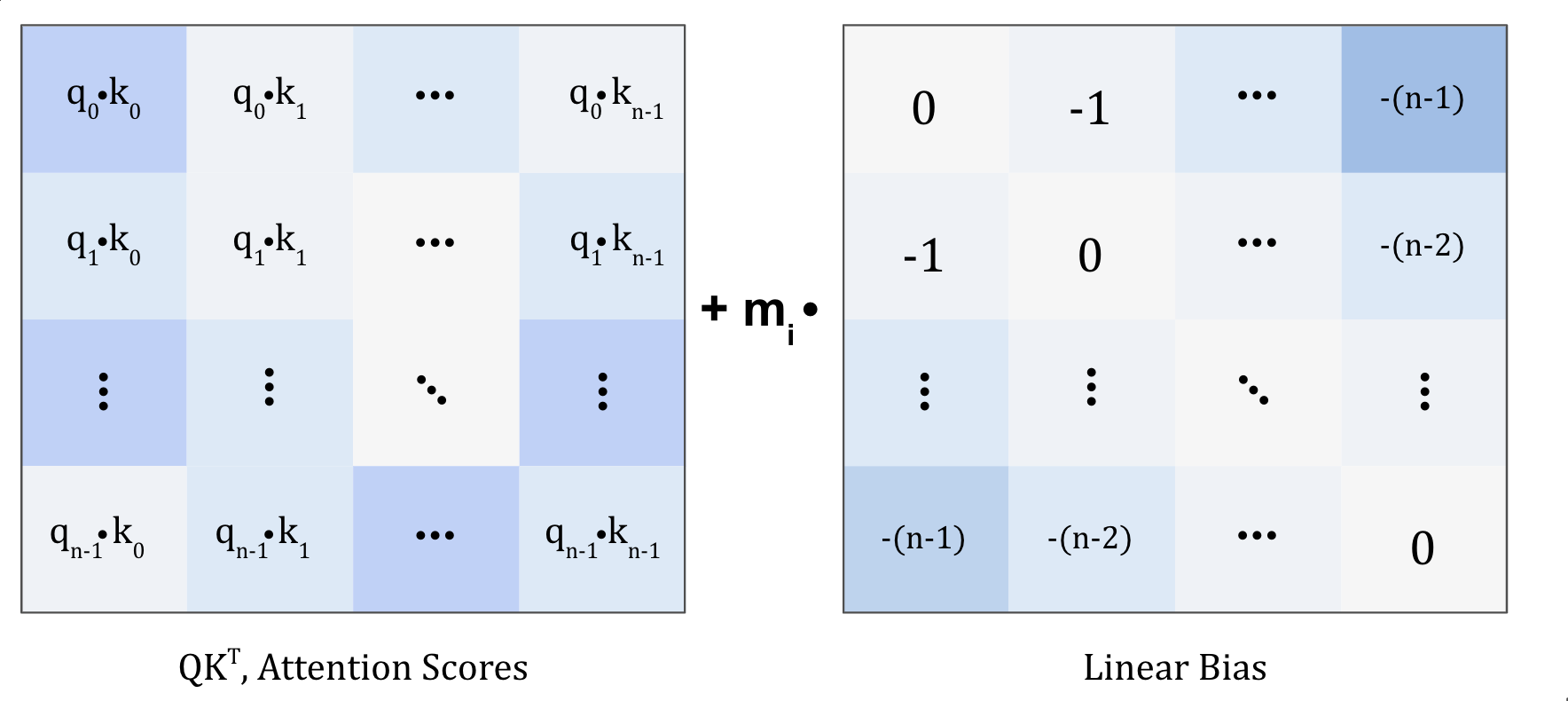}
    (a) Encoder ALiBi
\end{minipage}
\begin{minipage}[t]{0.5\textwidth}
    \centering
    \includegraphics[width=8cm]{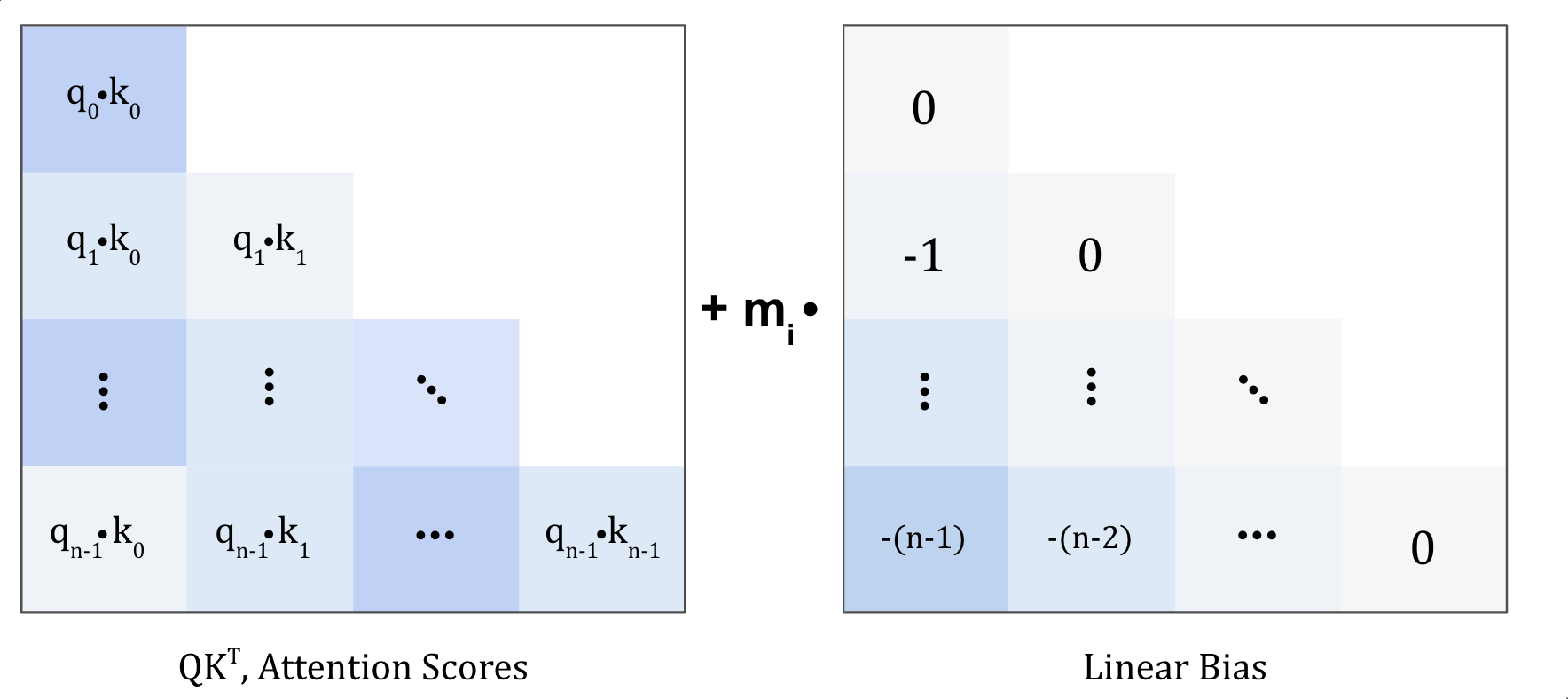}
    (b) Causal ALiBi
\end{minipage}
\caption[ddd]{With ALiBi attention, a linear bias is incorporated into each attention score preceding the softmax operation. Each attention head employs a distinct constant scalar, $m$, which diversifies its computation. Our model adopts the encoder variant where all tokens mutually attend during calculation, contrasting the causal variant originally designed for language modeling. In the latter, a causal mask confines tokens to attend solely to preceding tokens in the sequence.}
\label{fig:alibi-attention}
\end{figure*}

For the self-attention mechanism within the attention blocks, we adopt the Attention with Linear Biases (ALiBi) approach~\cite{press2022alibi}. ALiBi forgoes the use of positional embeddings. Instead, it encodes positional information directly within the self-attention layer by introducing a constant bias term to the attention score matrix of each layer, ensuring that proximate tokens demonstrate stronger mutual attention. While the original implementation was designed for causal language modeling and featured biases solely in the causal direction, such an approach is not compatible with the bidirectional self-attention inherent in our encoder model. For our purposes, we employ the symmetric encoder variant where attention biases are mirrored to ensure consistency in both directions\footnote{{\url{https://github.com/ofirpress/attention_with_linear_biases/issues/5}}}. Figure~\ref{fig:alibi-attention} depicts the computation of attention scores within the multi-head attention heads. Each head's scaling value, $m_i$, out of the total $n$ heads, is derived using Equation~\eqref{eq:alibi_bias}.

\begin{align}
\label{eq:alibi_bias}
&m_i = \begin{cases} 
      b ^ {2i} & i < a \\
      b ^ {1 + 2(i - a)} & i \ge a \\
   \end{cases} \nonumber\\
&a = 2^{\floor*{\log_2n}} \;\; b = 2 ^ \frac{-8}{2^{\ceil{\log_2n}}}
\end{align}

\paragraph{Gated Linear Units:}
For the feedforward sublayers within the attention blocks, we adopt Gated Linear Units (GLU), originally introduced in ~\cite{dauphin2016gatedconv}. They've demonstrated performance enhancements when incorporated into transformers~\cite{shazeer2020glu}. For the small and base models, we employ the GEGLU variant, which leverages the GELU activation function for the GLU. Conversely, for the large model, we utilize the ReGLU variant with the ReLU activation function. This choice was driven by our observation that training the large model with GEGLU, despite its promising initial MLM accuracy, was unstable.

\paragraph{Layer Normalization:}
Regarding Layer Normalization~\cite{ba2016layer}, we align with the post-layer normalization approach from \cite{vaswani2017transformer} in our attention blocks. Preliminary tests with pre-layer normalization, as mentioned in \cite{shoeybi2019megatron} and \cite{nguyen2019prenorm1}, didn't enhance training stability or performance. Consequently, we opted not to integrate it into our model.

\subsection{Training Data}
\label{sec:dataset:full-text}

For the pre-training phase, we leverage the English ``Colossal, Cleaned, Common Crawl (C4)'' dataset~\footnote{\url{https://huggingface.co/datasets/c4}}, encompassing approximately 365 million text documents harvested from the web, summing to around 170 billion tokens. As delineated in ~\cite{raffel2020exploring}, the C4 dataset is a refined iteration of Common Crawl, utilizing heuristics for cleanup and language recognition, retaining solely English content. As a result, our models are monolingual and tailored exclusively for English texts. The purification process also encompasses the removal of webpages hosting inappropriate content. We reserve $1\%$ of the dataset for evaluating validation loss and the accuracy of the masked language modeling (MLM) task.

\subsection{Training Algorithm}
Our model's pre-training revolves around the masked language modeling objective, excluding the next sentence prediction (NSP) task due to its perceived limited contribution to downstream task performance~\cite{liu2019roberta}. We mask $30\%$ of the input tokens randomly, employing whole word masking~\cite{devlin2019bert}, and condition the models to infer these masked tokens. Of these masked tokens, 80\% are substituted with the \texttt{[MASK]} token, $10\%$ with a random token, and the remaining $10\%$ stay unaltered.

The masked tokens are predicted by a decoder $f: \mathbb{R}^d \to \mathbb{R}^{|V|}$, which takes the output token embedding $\bm{e_i} \in \mathbb{R}^d$ of a masked token and predicts a probability for each token in the vocabulary. The loss $L_\mathrm{MLM}$ is computed by evaluating the cross entropy between the predicted probabilities and the actual masked tokens, as described in Equation~\eqref{eq:mlm}. Here, $I: \{1 ,\ldots, n\}\to|V|$ denotes the function that maps each of the $n$ masked tokens to its respective index in the vocabulary:
\begin{flalign}
\label{eq:mlm}
    \mathcal{L}_\mathrm{MLM}(t) &:= \sum\limits_{k=1}^n \ln f(\bm{e_i})_{I(k)}
\end{flalign}
Given our model's reliance on ALiBi attention~\cite{press2022alibi}, training position embeddings becomes unnecessary. This allows us to pre-train more efficiently on shorter sequences and adapt to longer sequences in subsequent tasks. Throughout our pre-training, we operate on sequences capped at $512$ tokens in length. Diverging from the methods in \cite{devlin2019bert} and \cite{liu2019roberta}, our sequences originate from individual documents without any multi-document packing. Furthermore, we refrain from sampling multiple sequences from a singular document. For each document, we exclusively consider its initial 512 tokens, truncating any excess. Given our consistent global batch size of 4096, each batch, due to its varying sequence length, contains a unique number of masked tokens when calculating loss.

\paragraph{Optimizer:}
Mirroring the optimization strategy of RoBERTa~\cite{liu2019roberta}, we employ the AdamW algorithm~\cite{loschilov2017adamw}, characterized by parameters $\beta_1 = 0.9$, $\beta_2 = 0.98$, $\epsilon = 1\mathrm{e}{-6}$, a weight decay of $0.01$, dropout set at $0.1$, and attention dropout also at $0.1$. Our learning rate schedule is linear, starting at $0$ and peaking at a rate of $\eta$ post $10,000$ steps. Here, the values of $\eta$ are designated as $1\mathrm{e}{-3}$, $6\mathrm{e}{-4}$, and $4\mathrm{e}{-4}$ for the small, base, and large models respectively. A linear decay to zero ensues after reaching the $100,000$ steps threshold.

\paragraph{Mixed precision training:} We resort to FP16 dynamic mixed precision~\cite{micikevicius2018mixed} for pre-training our models, facilitated by the DeepSpeed software package~\cite{rasley2020deepspeed}. Our preliminary tests using BF16 revealed unsatisfactory performance metrics, both in MLM accuracy and the downstream GLUE tasks.

\section{Fine-Tuning for Embeddings}
\label{sec:embedding-training}
After pre-training the \JBert~models, we further fine-tune each of the models to encode a text sequence into a single vector representation. The core idea behind our embedding approach is inspired by the Sentence-BERT~\cite{reimers2019sentence}. To enable a model to perform a text operation, we augment it with a mean pooling layer. This mean pooling step averages the token embeddings to merge their information into a single representation, without introducing additional trainable parameters. The training process for this enhanced model consists of an unsupervised phase followed by a supervised one.

\subsection{Fine-tuning with Text Pairs}
\label{subsec:unsupervised_pretraining}
During the first fine-tuning stage, we train the models on a corpus of text pairs $(q,p) \in \mathbb{D}^\mathrm{pairs}$, comprising a query string $q$ and a target string $p$.

\paragraph{Training Data}
We utilize roughly 40 diverse data sources, akin to the data preparation outlined in the report we previously published about our inaugural embedding model suite~\cite{gunther2023jina}. We observed that the inclusion of title-abstract pairs from documents significantly enhances performance on clustering tasks. As detailed in~\cite{gunther2023jina}, we implement consistency filtering~\citep{dai2023promptagator, wang2022text} to elevate the quality of the text pair corpus. For batch creation, we adhere to our earlier strategy: for every new batch, we randomly choose a data source and extract as many pairs as needed to fill that batch. All pairs within the data sources are pre-shuffled. Depending on the quality and quantity of the data sources, we assign different sampling rates for the pairs.

\paragraph{Loss Function:}
The goal of this fine-tuning stage is to encode text values that constitute a pair into analogous embedding representations, while encoding texts that aren't paired into distinct embeddings. To achieve this contrastive goal, we employ the InfoNCE~\citep{DBLP:journals/corr/abs-1807-03748} loss function, similar to our earlier embedding models~\cite{gunther2023jina}. This loss function calculates the loss value for a pair $(q,p) \sim \mathbf{B}$ within a batch $\mathbf{B} \subset \mathbb{D}^\mathrm{pairs}$ as follows:

\begin{align} 
    \mathcal{L}_{\nce}^{\mathrm{pairs}}(\mathbf{B}) := \mathbb{E}_{(q,p)\sim \mathbf{B}}\left[-\ln \frac{e^{s(q, p)/\tau}}{\sum\limits_{i = 1}^k e^{s(q, p_i)/ \tau}}\right]
\end{align}

The function evaluates the cosine similarity $s(p,q)$ between a given query $q$ and its corresponding target $p$, relative to the similarity of all other targets in the batch. Given the typically symmetric nature of similarity measures, we compute the loss in both directions:
\begin{flalign}
    \mathcal{L}^{\mathrm{pairs}}(\mathbf{B}) &:= \mathcal{L}^{\mathrm{pairs}}_{\nce}(\mathbf{B}) + \mathcal{L}^{\mathrm{pairs}}_{\overline{\nce}}(\mathbf{B}),\text{ with} \nonumber\\
    \mathcal{L}_{\overline{\nce}}^{\mathrm{pairs}}(\mathbf{B}) &:= \mathbb{E}_{(q,p)\sim \mathbf{B}}\left[-\ln \frac{e^{s(p, q) / \tau}}{\sum\limits_{i = 1}^k e^{s(p, q_i) / \tau}}\right]
\end{flalign}

The constant temperature parameter $\tau$ influences how the loss function weighs minor differences in the similarity scores~\cite{wang2021understanding}. Empirical testing suggests that $\tau=0.05$ is effective.

\subsection{Fine-tuning with Hard Negatives}
\label{sec:supervised_fine_tuning}

The goal of the supervised fine-tuning stage is to improve the models' ranking capabilities. This improvement is achieved by training with datasets that include additional negative examples.

\paragraph{Training Data} 
We have prepared retrieval datasets, such as MSMarco~\cite{bajaj2016ms} and Natural Questions (NQ)~\cite{47761}, in addition to multiple non-retrieval datasets like the Natural Language Inference (NLI) dataset~\cite{bowman2015large}. These datasets encompass a collection of queries with annotated relevant passages and several negative examples, consistent with earlier work~\cite{wang2022text}. Each training batch $B$, structured as $(q, p, n_1, \ldots, n_{15})$, includes one positive and 15 negative instances. For retrieval datasets, hard negatives are discerned by identifying passages deemed similar by retrieval models. This approach instructs the model to prioritize relevant documents over those that are merely semantically related. For non-retrieval datasets, negatives are selected randomly, since drawing a clear line between positives and hard negatives isn't feasible. This is because, unlike relevancy, it's challenging to make a binary determination regarding the similarity or dissimilarity of two textual values. Consequently, opting for hard negatives in such datasets seemed to diminish the models' quality. Nonetheless, it remains crucial to integrate these datasets into the stage~\ref{stage:three} training to ensure continued performance on non-retrieval tasks. To ensure that hard negative passages are indeed less relevant than the annotated relevant ones, we employ a cross-encoder model to validate that their relevance score is indeed lower.

\paragraph{Loss Function:} 
Our training employs a modified variant of the InfoNCE loss function, denoted as $\mathcal{L}_{\nce^+}$ and described by Equation \eqref{eq:loss-hard-negatives}. Similar to the preceding loss function, this one is bidirectional and incorporates the additional negatives when pairing queries with passages:
\begin{flalign}
&\mathcal{L}_{\nce^+}(B) := \nonumber\\
&\;\;\;\;\;\mathbb{E}_{r\sim B}\Bigg[-\ln \frac{e^{s(q, p) / \tau}}{\sum\limits_{i = 1}^k \Big[ e^{s(q, p_i) / \tau}+ \sum\limits_{j = 1}^{15} e^{s(q, n_{j,i}) / \tau}\Big]}\Bigg]\nonumber \\
&\, + \mathbb{E}_{r\sim B}\Bigg[-\ln \frac{e^{s(p, q) / \tau}}{\sum\limits_{i = 1}^k e^{s(p, q_i) / \tau}}\Bigg]\nonumber \\
&\text{with}\; r = (q,p, n_1, \ldots, n_{15}).\label{eq:loss-hard-negatives}
\end{flalign}

\subsection{Memory Optimizations}

When training embedding models, having a large batch size is crucial. This is because the InfoNCE loss functions $\mathcal{L}^\mathrm{pairs}$ and ${L}_{\nce^+}$ compute the loss values based on the entirety of the batch. The batch size determines the number of text values each individual text value is compared against. As a result, the computed loss value might not be as expressive with smaller batches. \citet{li2023general} provided an in-depth analysis, highlighting the positive impact of larger batch sizes on the performance of the resultant embedding model. To accommodate larger batch sizes, it becomes essential to minimize the memory overhead during training. We achieved this by training our models in mixed precision~\cite{micikevicius2018mixed} and leveraging the deepspeed~\cite{rasley2020deepspeed} framework for further optimization. Activation checkpointing~\cite{chen2016training} was also employed to curtail memory usage. Specifically, we inserted a checkpoint after each BERT layer within our model.

\section{Evaluation}
\label{sec:evaluation}

To evaluate the efficacy of our approach, we initiate with a comprehensive analysis of our pre-trained backbone models, as outlined in Section~\ref{sec:evaluation_jina_bert}. This is followed by an in-depth assessment of our embedding models in Section~\ref{sec:evaluation_jina_embedding}. Furthermore, we have conducted experiments to delve into the effects of encoding extended sequence lengths on the performance of the embeddings, presented in Section~\ref{sec:evaluation_seq_len}.

\subsection{Evaluation of \JBert}
\label{sec:evaluation_jina_bert}
%
%
\newcolumntype{P}[1]{>{\centering\arraybackslash}p{#1}}
\begin{table*}[htb]
    \centering
    \setlength{\tabcolsep}{4.5pt} 
\small{
\begin{tabular}{ l|rcccccccccc  }
 \toprule
 Model& Params & MNLI & QQP & QNLI & SST-2 & CoLa & STS-B & MRPC & RTE & WNLI & Average \\
 \midrule
BERT Base & 110M & 84.6/83.4 & 71.2 & 90.5 & 93.5 & 52.1 & 85.8 &  88.9 & 66.4 & - & - \\ 
BERT Large & 340M & 86.7/85.9 & 72.1 & 92.7 & 94.9 & 60.5 & 86.5 &   89.3 & 70.1 & -  & - \\ 
RoBERTa & 355M & 90.8/90.2 & 90.2 & 98.9 & 96.7 & 67.8 & 92.2 & 92.3 & 88.2 & 89.0 & 88.5 \\
    \midrule
\JBert~Small & 33M & 80.1/78.9 & 78.9 & 86.0 & 89.6 & 28.8 & 84.8 & 84.1 & 68.8 & 55.5 & 72.9 \\ 
\JBert~Base & 137M & 85.7/85.4 & 80.7 & 92.2 & 94.5 & 51.4  & 89.5 & 88.4 & 78.7 & 65.1 & 80.7 \\ 
\JBert~Large & 435M & 86.6/85.9 & 80.9 & 92.5 & 95.0 & 59.6  & 88.2 & 88.5 & 78.5 & 65.1 & 81.6 \\ 
\bottomrule
\end{tabular}
}
    \caption{Evaluation of the \JBert~models on the GLUE benchmark} 
    \label{tab:glue_results}
\end{table*}
\begin{figure}[htbp!]
  \centering
  \includegraphics[width=\linewidth]{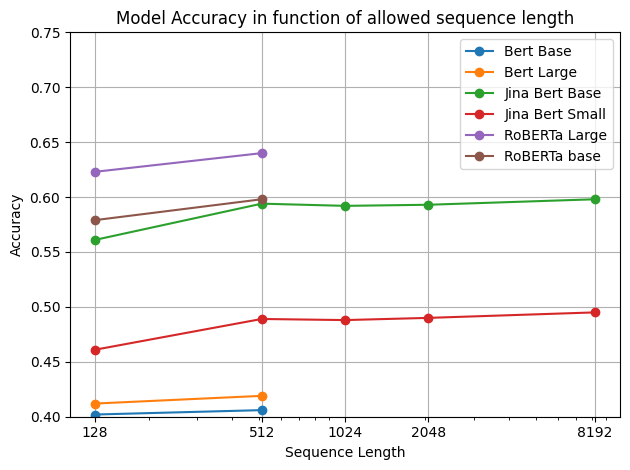}
  \caption{Variation of model MLM accuracy w.r.t. the sequence length}
  \label{fig:long_doc_accuracy}
\end{figure}

Following previous work \cite{DBLP:journals/corr/abs-1907-11692}, we evaluate our pretrained models on the GLUE benchmark \cite{DBLP:journals/corr/abs-1804-07461}. General Language Understanding Evaluation (GLUE) is a collection of nine datasets for evaluating natural language understanding systems.
Six tasks are framed as either single-sentence classification or sentence-pair classification tasks.
The GLUE organizers provide training, development, and test data splits, as well as a submission server and leaderboard.\footnote{\url{https://gluebenchmark.com}}
The test split does not contain labels, and the submission server allows participants to evaluate and compare their systems against the private labels of the test split.

For the \JBert~training described in Section~\ref{sec:backbone_pretraining}, we fine-tune the pre-trained models on the corresponding single-task training data using several hyperparameter settings and, for each task, pick the best fine-tuning hyperparameters on the development set.

Following the methodology of \cite{phang2018mnlitrick}, for RTE, STS, and MRPC, we fine-tune starting from the MNLI single-task model, rather than the baseline pretrained \JBert~models.
As in the BERT paper~\cite{devlin2019bert}, our fine-tuning procedure relies on representing the input sequence and using the final hidden vector \(C \in \mathbb{R}^ H\) corresponding to the first input token (\texttt{[CLS]}) as the aggregate representation.

We train for 10 epochs with batch sizes $\{16, 32\}$ and learning rates $\{1\mathrm{e}{-5}, 2\mathrm{e}{-5}, 3\mathrm{e}{-5}\}$. For each task, the best fine-tuned model on the development set is used for the test set.

In Table \ref{tab:glue_results}, we report the results of the best-performing models on the test sets after submission to the GLUE benchmark server.

Furthermore, we evaluate \JBert~models on documents of long text sequences by computing the accuracy of the MLM task with varying sequence lengths. The accuracy of masked language modeling is computed on $50,000$ samples from the C4 validation set where, for each chosen sequence length, each sample document is tokenized and truncated to fit the sequence length.
We compare \JBert~to RoBERTa and BERT models in Figure~\ref{fig:long_doc_accuracy}. It essentially shows that, even though \JBert~models were trained on a $512$ sequence length, the MLM accuracy does not drop when we extrapolate to an $8192$ sequence length. For other BERT and RoBERTa models, since they use absolute positional embeddings that are trained on a $512$ sequence length, it's not possible to compute the MLM accuracy beyond $512$. The figure demonstrates ALiBi's effectiveness in maintaining MLM performance during inference for long documents.

\subsection{Evaluation of \JEmbeddingVTwo}
\label{sec:evaluation_jina_embedding}

To comprehensively evaluate our embedding models, we employ the Massive Text Embedding Benchmark (MTEB)~\cite{muennighoff2023mteb}. Our choice of MTEB is motivated by its unparalleled breadth, distinguishing it among embedding benchmarks. Rather than focusing on a single task and dataset, MTEB covers an expansive set of 8 tasks, encompassing a rich collection of 58 datasets across 112 languages. This expansive benchmark allows us to scrutinize our model's adaptability across diverse applications and languages and benchmark it against other top-performing models. 

However, a limitation of the MTEB benchmark is its omission of very long texts, which are essential for evaluating our model's prowess in handling $8192$ sequence lengths. Consequently, we introduce new retrieval and clustering tasks featuring extended documents, and we detail the performance of our model against its peers in Section \ref{sec:evaluation_seq_len}.

\textbf{Clustering}: The goal here is to aptly group a collection of sentences or paragraphs. Within the MTEB benchmark suite, a mini-batch $k$-means model is employed, operating with a batch size of 32. Here, $k$ represents the number of unique labels in the dataset. Model performance is evaluated using the $\mathcal{V}$ measure, a metric insensitive to cluster label permutations, guaranteeing that assessments are independent of label configurations.

We incorporate two new clustering tasks featuring extended documents within the MTEB clustering task subset. The inaugural task, named PatentClustering, draws from the BigPatent\footnote{\url{https://huggingface.co/datasets/big_patent}} dataset~\cite{DBLP:journals/corr/abs-1906-03741}, challenging the k-means model to organize patents by their respective categories. Patent documents average $6,376$ tokens, spanning a range from a brief $569$ tokens to an extensive $218,434$ tokens. Our second task, titled WikiCitiesClustering, sources from the English subset of the refined Wikipedia dump \cite{wikidump}, available as a dataset on Hugging Face\footnote{\url{https://huggingface.co/datasets/wikipedia}}. For this task, we curate a roster of nations from \href{https://www.wikidata.org/wiki/Wikidata:Main_Page}{Wikidata} and extract Wikipedia articles of their cities from the refined dataset. The objective is to group cities by their parent country. On average, articles consist of $2,031$ tokens, with the length varying between a succinct 21 tokens to a comprehensive $20,179$ tokens.

\textbf{Retrieval}: This task entails a dataset comprising a corpus, a set of queries, and associated mappings connecting each query to pertinent corpus documents. The mission is to discern relevant documents for a specific query. Both queries and corpus documents undergo encoding, after which their similarity scores are derived using cosine similarity. Subsequently, metrics like nDCG$@10$ (which serves as the primary metric), MRR$@k$, MAP$@k$, precision$@k$, and recall$@k$ are computed for diverse $k$ values. This task is inspired by datasets and evaluation methods presented by BEIR \cite{thakur2021beir}. 

To expand the scope of the MTEB, we introduce a new retrieval task named NarrativeQA, derived from the \texttt{narrativeqa}\footnote{\url{https://huggingface.co/datasets/narrativeqa}} dataset. This dataset boasts realistic QA instances, curated from literature (encompassing both fiction and non-fiction) and film scripts. The corpus averages $74,843$ tokens per document, with the lengthiest document tallying up to $454,746$ tokens, and the most concise one comprising $4,550$ tokens.

We further evaluated \JEmbeddingVTwo~using a novel benchmark, referred to as LoCo~\footnote{\url{https://hazyresearch.stanford.edu/blog/2024-01-11-m2-bert-retrieval}}. The LoCo dataset consists of five retrieval tasks derived from publicly available datasets. The selection process for these tasks was guided by several criteria, notably the length of the documents, with a preference towards longer texts, in addition to a manual review to verify that the tasks require a thorough understanding of the entire document. The results of our models on the LoCo dataset are provided in Table~\ref{tab:locoresult}.

\subsubsection{Results on MTEB}

\begin{table*}[htb]
    \centering
    \setlength{\tabcolsep}{4.5pt} 
\small{
\begin{tabular}{ l|rccccccccc  }
 \toprule
 Model& Params & CF & CL & PC & RR & RT & STS & SM & Average \\
 \midrule
\AdaTwo & unknown & 70.93 & 45.90 & 84.89 & 56.32 & 49.25 & 80.97 & 30.80 & 60.99 \\
e5-base-v2 & 110M & 73.84 & 43.80 & 85.73 & 55.91 & 50.29 & 81.05 & 30.28 & 61.50 \\
all-MiniLM-L6-v2 & 23M & 63.05 & 42.35 & 82.37 & 58.04 & 41.95 & 78.90 & 30.81 & 56.26 \\
all-mpnet-base-v2 & 110M & 65.07 & 43.69 & 83.04 & 59.36 & 43.81 & 80.28 & 27.49 & 57.78 \\
    \midrule
\JinaSTwo & 33M & 68.82 & 40.08 & 84.44 & 55.09 & 45.64 & 80.00 & 30.56 & 58.12 \\
\JinaBTwo & 137M & 73.45 & 41.74 & 85.38 & 56.99 & 48.45 & 80.70 & 31.60 & 60.37 \\
 \bottomrule
\end{tabular}

CF: Classification Accuracy [\%] \quad{}
CL: Clustering $\mathcal{V}$ measure[\%]\quad{}
PC: Pair Classification Average Precision [\%]\quad{} \\
RR: Reranking MAP [\%]\quad{}
RT: Retrieval nDCG@10\quad{}
STS: Sentence Similarity Spearman Correlation [\%]\quad{} \\
SM: Summarization Spearman Correlation [\%]\quad{}
}
    \caption{Evaluation of the \JEmbeddingVTwo~models on the MTEB benchmark} 
    \label{tab:mteb_results}
\end{table*}
\begin{figure*}[htb!]
\begin{minipage}[t]{0.5\textwidth}
    \centering
    \includegraphics[width=8cm]{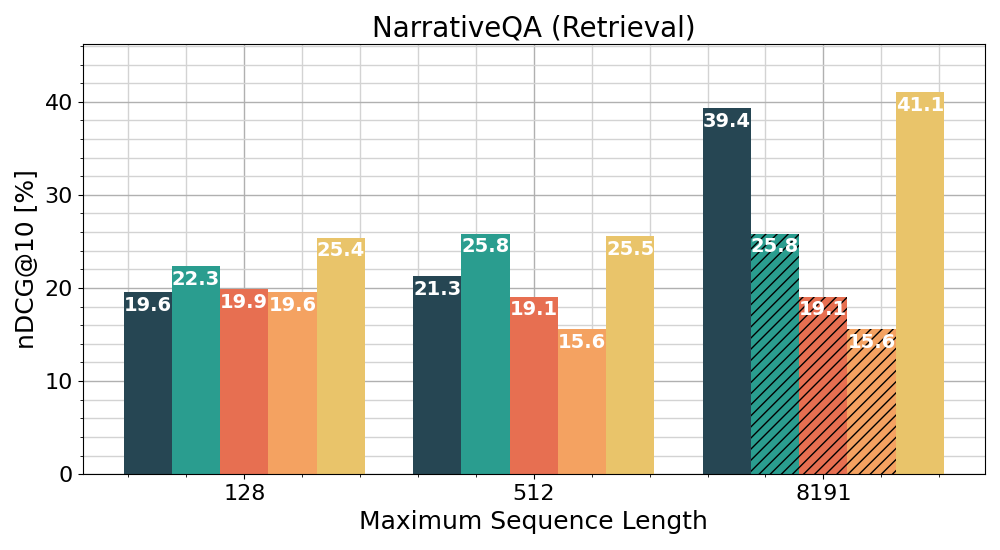}
\end{minipage}
\begin{minipage}[t]{0.5\textwidth}
    \centering
    \includegraphics[width=8cm]{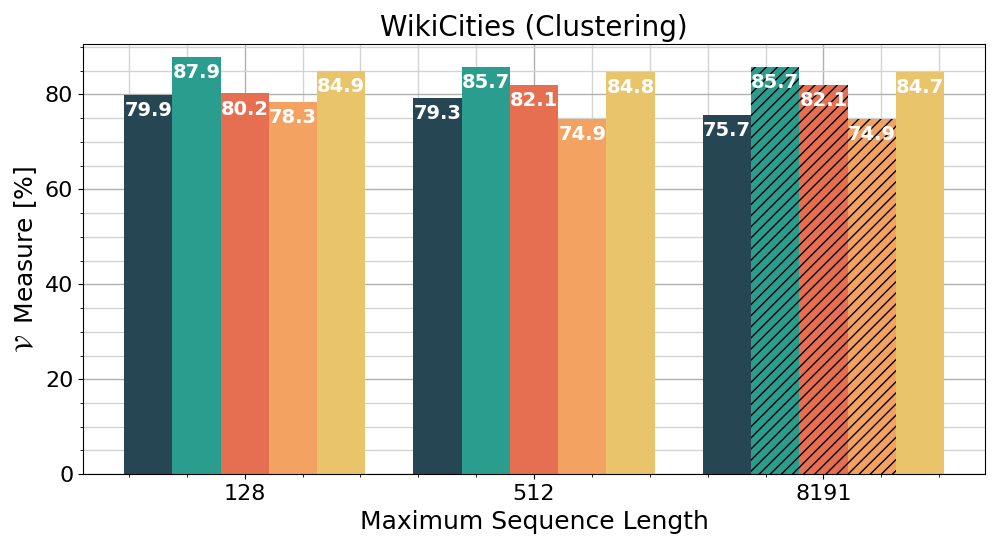}
\end{minipage}
\begin{minipage}[t]{0.5\textwidth}
    \centering
    \includegraphics[width=8cm]{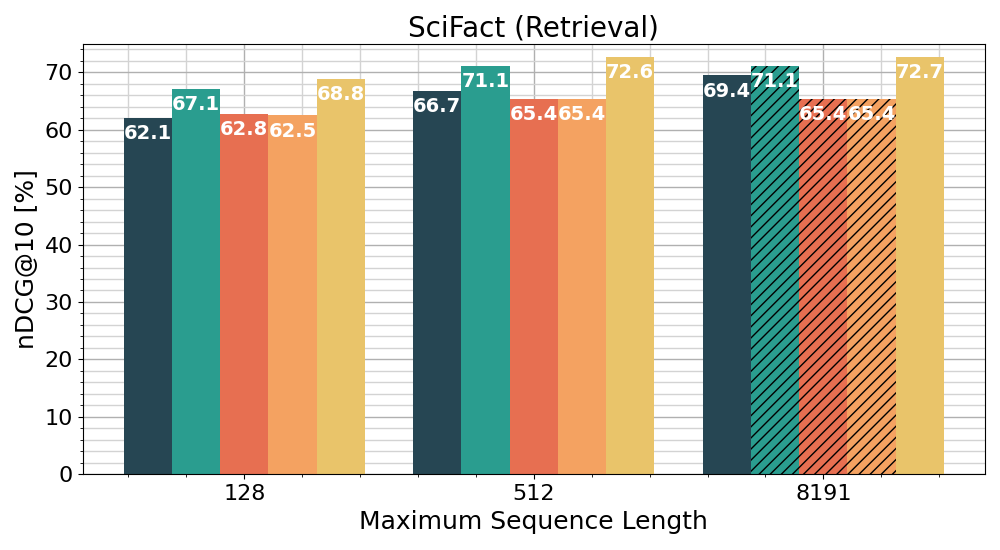}
\end{minipage}
\begin{minipage}[t]{0.5\textwidth}
    \centering
    \includegraphics[width=8cm]{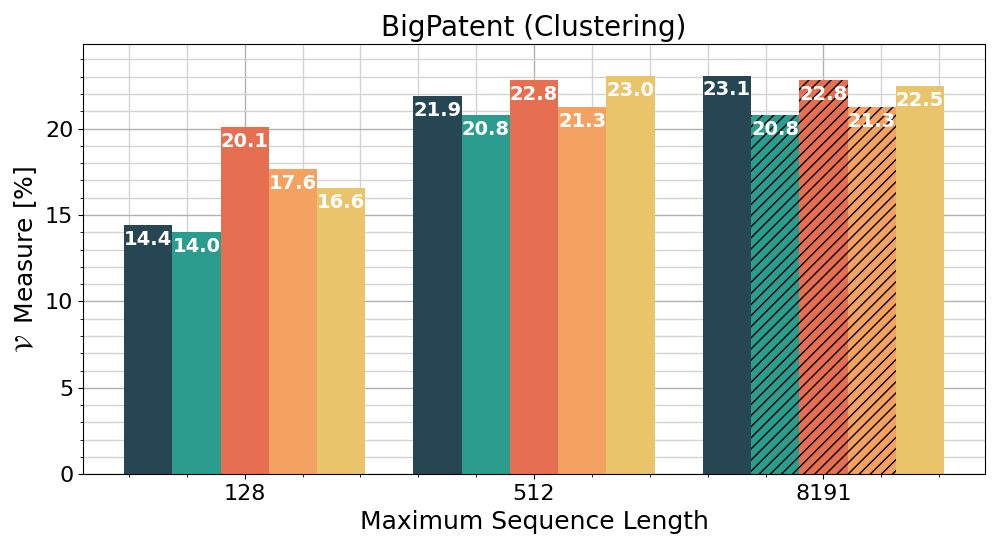}
\end{minipage}
\begin{minipage}[t]{\textwidth}
    \centering
    \includegraphics[width=14cm]{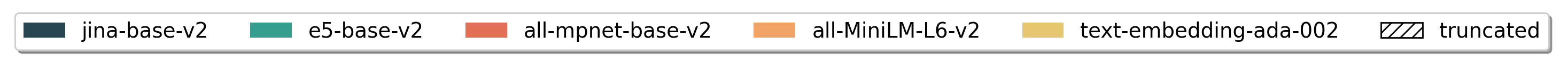}
\end{minipage}
\caption[Evaluation with different maximum sequence lengths]{Evaluation w.r.t. maximum sequence length. For \texttt{e5-base-v2}, we abstained from employing specific prefixes like \texttt{query: }, which might result in varied evaluation outcomes. Note, \texttt{text-embedding-ada-002} caps its context length at $8191$ tokens, not $8192$.}
\label{fig:eval-seq-len}
\end{figure*}

The evaluation of embedding models within the MTEB benchmark, as illustrated in Table \ref{tab:mteb_results}, reveals significant contrasts between Jina's text embedding models, namely \JinaSTwo~and \JinaBTwo, and other contemporary models. These differences are especially pronounced in tasks showing marked performance disparities, such as Classification (CF) and Retrieval (RT).

In Classification (CF), the \JinaBTwo{} model, equipped with 137 million parameters, emerges as a leading performer. It records superior scores, outpacing most competing models, underscoring its efficacy in text classification. Conversely, the \JinaSTwo{} model, equipped with a modest 33 million parameters, trails behind some other models in this task. This underscores the pivotal role model size plays in certain downstream tasks, with more extensive architectures yielding potential benefits.

For the Retrieval (RT) task, \JinaSTwo{} showcases formidable performance, signaling its adeptness for information retrieval. It ranks amidst top-tier models, indicating its prowess in retrieval-centric tasks. Similarly, \JinaBTwo{} excels, registering a slightly superior score, reaffirming its formidable retrieval aptitude. Both models underscore their credibility in tasks necessitating adept information retrieval. Given that models \texttt{all-MiniLM-L6-v2} and \texttt{all-mpnet-base-v2} omit the second-stage finetuning which \JinaSTwo{} and \JinaBTwo{} undergo, it's foreseeable that our models would excel in these tasks.

In conclusion, both the base and small text embedding models display commendable performance within the MTEB benchmark. Their standout performance, relative to other models in tasks like Classification and Retrieval, suggests model size's influential role in specific text processing endeavors. Both models reaffirm their potency in retrieval, marking them as pivotal tools for a plethora of natural language processing tasks.

\subsubsection{Impact of Maximum Sequence Length}
\label{sec:evaluation_seq_len}
As delineated in Section~\ref{sec:evaluation_jina_bert}, the pre-training generalizes across extended sequence lengths. Consequently, the MLM accuracy for long sequences, spanning up to $8192$ tokens, mirrors that of shorter sequences, despite the exclusive training on abbreviated text sequences. During finetuning, our models train solely on texts not exceeding $512$ tokens, yet they cater to texts reaching $8192$ tokens for the MTEB evaluation detailed in Section~\ref{sec:evaluation_jina_embedding}.

To discern how sequence length impacts the accuracy of downstream tasks, we executed long document clustering and retrieval tasks, modulating the tokenizer's maximum sequence length. This allows us to gauge the models' performance on variable sequence lengths through truncation. Since a majority of the extant tasks in the MTEB feature documents under $512$ tokens, we resort to our three novel datasets elucidated in Section \ref{sec:evaluation_jina_embedding}, accessible on Hugging Face. Furthermore, we employ the SciFact dataset~\cite{wadden2020fact}, given its substantial count of texts exceeding $512$ tokens.

Figure~\ref{fig:eval-seq-len} depicts the nDCG$@10$ retrieval and the $\mathcal{V}$ measure scores for the \JinaBTwo{} alongside four other renowned embedding models. Given that only \JinaBTwo{} and OpenAI's \texttt{text-embedding-ada-002} support an 8K sequence length, results reported for an 8191 sequence length for other models are truncated to their intrinsic maximum, typically $512$. Generally, Figure~\ref{fig:eval-seq-len} suggests that elongated sequence lengths contribute to enhanced outcomes. This assertion is particularly true for the NarrativeQA task, where extending the sequence length substantially bolsters performance. Due to the inherent nature of the dataset, models limited to the text's commencement frequently underperform.

On the BigPatent clustering task, larger sequence lengths also result in better performance. However, on the WikiCities clustering task, longer sequence lengths seem to slightly diminish the models' performance in most instances. This suggests that an increase in sequence length doesn't always yield better outcomes. One explanation for this observation is that the initial paragraph of a Wikipedia article about a city typically mentions the country the city is in. Information towards the middle and end of the articles is often less pertinent for identifying the country and might alter the attributes that influence the clustering of the city embeddings. 

\section{Conclusion}
\label{sec:conclusion}
We have introduced \JEmbeddingVTwo, a novel embedding model based on a modified BERT architecture. This model eschews positional embeddings and instead employs bi-directional ALiBi slopes to capture positional information. By training a series of embedding models with this innovative architecture on the Web document corpus C4 and subsequently fine-tuning them, we have enabled the encoding of the semantics of both short and long textual values into meaningful vector representations. This effort has produced a new suite of open-source embedding models capable of encoding texts containing up to $8192$ tokens. These embeddings signify a 16x increase in the maximum sequence length compared to leading open-source embedding models. Additionally, our model suite exhibits competitive performance on the MTEB benchmark. We also demonstrate how utilizing extended sequence lengths can offer our models an advantage over those without such capabilities.

\balance

\bibliographystyle{unsrtnat}
\bibliography{references}  

\clearpage
\pagenumbering{gobble}
\onecolumn

\appendix
\section{Appendix: MTEB and LoCo Becnharmk}
\centering
\begin{table*}[ht]
\centering
\begin{tabular}{ l|rr  }
 \toprule
 & \multicolumn{2}{c}{ Accuracy [\%]} \\
 Task& \JinaSTwo & \JinaBTwo  \\
 \midrule
 AmazonCounterfactualClassification& 71.36 & 74.73 \\
 AmazonPolarityClassification& 82.90 & 88.54 \\
 AmazonReviewsClassification& 40.89 & 45.26  \\
 Banking77Classification & 78.25 & 84.01 \\
 EmotionClassification & 44.01 & 48.77 \\
 ImdbClassification & 73.64 & 79.44 \\
 MassiveIntentClassification & 67.61 & 71.93 \\
 MassiveScenarioClassification & 69.75 & 74.49 \\
 MTOPDomainClassification & 93.96  & 95.68 \\
 MTOPIntentClassification & 72.50 & 83.15 \\
 ToxicConversationsClassification & 71.54 & 73.35 \\
 TweetSentimentExtractionClassification & 59.40 & 62.06 \\
 \midrule
 Avg & 68.82 & 73.45 \\
 \bottomrule
\end{tabular}
\caption{Detailed Performance on the MTEB Classification Tasks}
\end{table*}
\centering
\begin{table*}[ht]
\centering
\begin{tabular}{ l|rr  }
 \toprule
 & \multicolumn{2}{c}{$\mathcal{V}$ measure} \\
 Task& \JinaSTwo & \JinaBTwo  \\
 \midrule
 ArxivClusteringP2P& 44.02 & 45.39 \\
 ArxivClusteringS2S& 35.16 & 36.68 \\
 BiorxivClusteringP2P& 35.57 & 37.05  \\
 BiorxivClusteringS2S & 29.07 & 30.16 \\
 MedrxivClusteringP2P & 31.86 & 32.41 \\
 MedrxivClusteringS2S & 27.51 & 28.09 \\
 RedditClustering & 49.28 & 53.05 \\
 RedditClusteringP2P & 57.09 & 60.31 \\
 StackExchangeClustering & 55.35  & 58.52 \\
 StackExchangeClusteringP2P & 34.42 & 34.96 \\
 TwentyNewsgroupsClustering & 41.57 & 42.47 \\
 \midrule
 Avg & 40.08 & 41.73 \\
 \bottomrule
\end{tabular}
\caption{Detailed Performance on the MTEB Clustering Tasks}
\end{table*}
\centering
\begin{table*}[ht]
\centering
\begin{tabular}{ l|rr }
 \toprule
 & \multicolumn{2}{c}{Spearman correlation based on $\cos$ similarity} \\
 
 Task& \JinaSTwo & \JinaBTwo  \\
 \midrule
 SummEval& 30.56 & 31.60 \\
 \bottomrule
\end{tabular}
\caption{Detailed Performance on the MTEB Summarization Tasks}
\end{table*}
\centering
\begin{table*}[ht]
\centering
\begin{tabular}{ l|rr  }
 \toprule
 & \multicolumn{2}{c}{$\cos$\texttt{-sim-ap}} \\

 Task& \JinaSTwo & \JinaBTwo  \\
 \midrule
 SprintDuplicateQuestions& 95.12 & 95.30 \\
 TwitterSemEval2015& 72.15 & 74.74 \\
 TwitterURLCorpus& 86.05 & 86.09  \\
 \midrule
 Avg & 84.44 & 85.38 \\
 \bottomrule
\end{tabular}
\caption{Detailed Performance on the MTEB Pair Classification Tasks}
\end{table*}
\centering
\begin{table*}[ht]
\centering
\begin{tabular}{ l|rr  }
 \toprule
 & \multicolumn{2}{c}{mAP@10} \\
 Task& \JinaSTwo & \JinaBTwo  \\
 \midrule
 AskUbuntuDupQuestions& 59.62 & 62.25 \\
 MindSmallReranking& 30.99 & 30.54 \\
 SciDocsRR& 79.76 & 83.10  \\
 StackOverflowDupQuestions& 49.99 & 52.05  \\
 \midrule
 Avg & 55.09 & 56.98 \\
 \bottomrule
\end{tabular}
\caption{Detailed Performance on the MTEB ReRanking Tasks}
\end{table*}
\centering
\begin{table*}[ht]
\centering
\begin{tabular}{ l|rr }
 \toprule
 & \multicolumn{2}{c}{nDCG@10} \\
 
 Task& \JinaSTwo & \JinaBTwo  \\
 \midrule
 ArguAna& 46.73 & 44.18 \\
 ClimateFEVER& 20.05 & 23.53 \\
 CQADupstackRetrieval& 38.03 & 39.34  \\
 DBPedia& 32.65 & 35.05  \\
 FEVER& 68.02 & 72.33 \\
 FiQA2018& 33.43 & 41.58 \\
 HotpotQA& 56.48 & 61.38  \\
 MSMARCO& 37.28 & 40.92  \\
 NFCorpus& 30.40 & 32.45  \\
 NQ& 51.59 & 60.44 \\
 QuoraRetrieval& 87.19 & 88.20  \\
 SCIDOCS& 18.61 & 19.86  \\
 SciFact& 63.89 & 66.68  \\
 Touche2020& 23.52 & 26.24 \\
 TRECCOVID& 65.18 & 65.91  \\
 \midrule
 Avg & 45.14 & 47.87 \\
 \bottomrule
\end{tabular}
\caption{Detailed Performance on the MTEB Retrieval Tasks}
\end{table*}
\centering
\begin{table*}[ht]
\centering
\begin{tabular}{ l|rr  }
 \toprule
 & \multicolumn{2}{c}{Spearman correlation based on cosine similarity} \\
 
 Task& \JinaSTwo & \JinaBTwo  \\
 \midrule
 BIOSSES& 80.52 & 81.23 \\
 SICK-R& 76.72 & 79.65 \\
 STS12& 73.66 & 74.27  \\
 STS13& 83.30 & 84.18  \\
 STS14& 79.17 & 78.81 \\
 STS15& 87.30 & 87.55 \\
 STS16& 83.61 & 85.35  \\
 STS17(en-en)& 88.23 & 88.88  \\
 STS22(en)& 63.46 & 62.20  \\
 STSBenchmark& 84.04 & 84.84 \\
 \midrule
 Avg & 80.00 & 80.70 \\
 \bottomrule
\end{tabular}
\caption{Detailed Performance on the MTEB STS Tasks}
\end{table*}

\centering
\begin{table*}[ht]
\centering
\begin{tabular}{ l|rrrr  }
\toprule
 Model& Fine-tuned on LoCo & Parameters & Context Length & avg. nDCG@10 \\
 \midrule
 M2-BERT-32768 & \checkmark & 80M & 32,768 &  92.5 \\
 e5-mistral-7b-instruct &  & 7.3B & 4,096 &  88.5 \\
 M2-BERT-32768 & \checkmark & 80M & 8,192 &  85.9 \\
 \JinaBTwo &  & 137M & 8192 &  85.4 \\
 bge-large-en-v1.5 & \checkmark & 335M & 512 &  85.0 \\
 M2-BERT-2048 & \checkmark & 80M & 2,048 &  83.6 \\
 \JinaSTwo &  & 33M & 8,192 &  83.4 \\
 bge-base-en-v1.5 & \checkmark & 109M & 512 &  83.0 \\
 bge-small-en-v1.5 & \checkmark & 33M & 512 &  81.2 \\
 bge-large-en-v1.5 &  & 335M & 512 &  77.2 \\
 bge-base-en-v1.5 &  & 109M & 512 &  73.4 \\
 bge-small-en-v1.5 &  & 33M & 512 &  70.6 \\
 cohere-embed-v3 &  & NA & 512 &  66.6 \\
 ada-embeddings-002 &  & NA & 8,191 &  52.7 \\
 voyage-v1 &  & NA & 4,096 &  25.4 \\
 \bottomrule
\end{tabular}
\caption{Performance on the new LoCo Dataset\label{tab:locoresult}}
\end{table*}

\end{document}